\documentclass{article}
\usepackage[preprint]{colm2026_conference}
\usepackage[utf8]{inputenc}
\usepackage{microtype}
\usepackage{natbib}
\usepackage{hyperref}
\usepackage{url}
\usepackage{booktabs}
\usepackage{graphicx}
\usepackage{amsmath}
\usepackage{amssymb}
\usepackage{adjustbox}
\usepackage{subcaption}
\usepackage{xcolor}
\usepackage{tcolorbox}

\title{Don't Make the LLM Read the Graph: Make the Graph Think\\
\large Architectural Integration of Belief Graphs for Multi-Agent LLM Reasoning}

\author{%
  Yuqi Sun \\
  Mindoverflow\\
  \And
  Tianqin Meng \\
  University of Waterloo \\
  \And
  George Liu  \\
  Carnegie Mellon University \\
  \And
   Yashraj Panwar \\
  Foothill College \\
  \And
  Lakshya Chaudhry  \\
  Purdue University \\
  \And
  Munasib Ilham \\
  University of Wisconsin \\
  \And
  Aman Chadha\\
  Apple\thanks{Work done outside role at Apple.} \\
}

\begin{document}
\maketitle

\begin{abstract}
We investigate whether explicit belief graphs improve LLM performance in cooperative multi-agent reasoning. Through 3,000+ controlled trials across four LLM families in the cooperative card game Hanabi, we establish four findings. \textbf{First}, integration architecture determines whether belief graphs provide value: as prompt context, graphs are decorative for strong models and beneficial only for weak models on 2nd-order Theory of Mind (80\% vs 10\%, p$<$0.0001, OR=36.0); when graphs gate action selection through ranked shortlists, they become structurally essential even for strong models (100\% vs 20\% on 2nd-order ToM, p$<$0.001). \textbf{Second}, we identify ``Planner Defiance,'' a model-family-specific failure where LLMs override correct planner recommendations at partial competence (90\% override, replicated N=20); Gemini models show near-zero defiance while Llama 70B shows 90\%, and models distinguish factual context (deferred to) from advisory recommendations (overridden). \textbf{Third}, full-game evidence confirms inter-agent conventions (+128\% over baseline, p=0.003) outperform all single-agent interventions, and individual belief-graph components must be combined to produce gains. \textbf{Fourth}, preliminary scaling analysis (N=10/cell, exploratory) suggests graph depth has diminishing returns: shallow graphs provide the best cost-benefit ratio, while deeper ToM graphs appear harmful at larger player counts ($-$1.5 pts at 5-player, p=0.029).
\end{abstract}

\section{Introduction}

Large Language Models have demonstrated strong reasoning capabilities, yet their performance in multi-agent cooperative settings remains limited. A key challenge is Theory of Mind (ToM), which involves modeling what other agents know, believe, and plan. In cooperative games like Hanabi, correct play requires multi-step inference about partners' knowledge states.

A natural approach is to provide LLMs with explicit \textit{belief graphs}: structured representations of agent knowledge. But a fundamental question remains: \textbf{does the belief graph help because it provides information the LLM lacks (scaffolding), or because it performs computation the LLM cannot do (planning)?}

This distinction matters for the future of tool-augmented AI. If belief graphs are scaffolding, they will become obsolete as models improve. If they perform irreplaceable computation, namely forward-simulating belief updates and searching action consequences in belief space, they provide permanent orthogonal value.

We present a systematic investigation using six diagnostic scenarios that each isolate a specific ToM competency (Table~\ref{tab:t1}), three integration architectures, and four graph-ablation conditions. Our four contributions are:

\begin{enumerate}
\item \textbf{Architecture determines graph value.} As prompt context, graphs are decorative for strong models and beneficial only for weak models on 2nd-order ToM: 80\% vs 10\% with full vs absent graph (N=20, p$<$0.0001, OR=36.0); all other scenarios are at ceiling, confirmed by a concurrent null control (p=1.0). But when the graph instead \textit{constrains} the LLM's choices to a ranked shortlist of actions computed via belief-space search (\S\ref{sec:architectures}), accuracy reaches 100\% on the same task where all prompt-based approaches score $\leq$20\% (p$<$0.001), replicated on Llama 70B (96.7\% vs 86.7\%). This establishes the information-pipeline vs.\ decision-pipeline distinction.

\item \textbf{Planner Defiance.} When given access to the belief graph's action recommendations but allowed to override them, LLMs reject correct suggestions specifically at partial competence (90\% override rate, N=20, p$<$0.0001). Defiance is \textit{model-family-specific}: Gemini families show 0 to 5\% override; Llama 70B shows 90\%. Models defer to factual context but override advisory recommendations. A candidate mitigation through explanatory reasoning requires replication at N$\geq$50.

\item \textbf{Conventions and system integration.} Inter-agent conventions are the only intervention that significantly improves cooperative play (6.4/25 vs baseline 2.8/25, p=0.003). Individual belief-graph components must be combined to produce benefits, and the graph's primary role in full games is state extraction rather than reasoning.

\item \textbf{Preliminary scaling sensitivity.} Exploratory analysis (N=10/cell, single model family) suggests graph depth has diminishing returns: shallow graphs provide the best cost-benefit ratio at 3-player ($+$1.0 pts, p=0.061, not significant), while deeper ToM graphs appear harmful at 5-player ($-$1.5 pts, p=0.029). Graph depth should be calibrated to environmental complexity rather than maximized.
\end{enumerate}

\section{Related Work}

\textbf{Theory of Mind in LLMs.} Recent work evaluating LLMs on ToM benchmarks yields mixed results: frontier models pass basic false-belief tasks but struggle with higher-order reasoning \citep{kosinski2024,shapira2023}. Our work extends this to an interactive game setting requiring multi-step inference under imperfect information.

\textbf{LLMs in cooperative games.} Hanabi has served as a multi-agent AI testbed, with rule-based and RL agents achieving near-perfect scores \citep{bard2020}. Prior LLM+Hanabi work typically uses prompt-only configurations and evaluates via aggregate game scores, without ablating why belief graphs help or hurt. We systematically compare three architectures and four ablation conditions, using diagnostic scenarios to isolate which ToM demands are graph-sensitive.

\textbf{External tools and LLM reasoning.} Integrating external tools with LLMs has shown significant benefits \citep{schick2023}. Our belief-space planner computes action consequences through forward simulation rather than providing facts, connecting to verifiers \citep{cobbe2021} and process reward models \citep{lightman2024}. The distinction we formalize between information-providing and computation-providing tools predicts that only the latter retains value as model capability grows. We also connect to information overload work \citep{liu2024}: our full-game results show richer context can increase recklessness.

\textbf{Unique position.} Four contributions absent from prior LLM+Hanabi work: (1) controlled ablation isolating graph quality from architecture, establishing the information-pipeline vs.\ decision-pipeline distinction; (2) Planner Defiance, a model-family-specific failure mode visible only in the hybrid architecture; (3) evidence that conventions dominate all single-agent interventions, with component synergy analysis; and (4) the first multi-player scaling analysis providing preliminary evidence that graph depth should be calibrated to player count. See Appendix~\ref{app:usps} for a direct comparison.

\section{Methodology}

\subsection{Hanabi as a ToM Testbed}

Hanabi is a 2 to 5 player cooperative card game where each player sees all cards except their own. Players share limited hint tokens to communicate card identities. Correct play requires 1st-order ToM, 2nd-order ToM, cooperative intent inference, and forward simulation.

Rather than measuring aggregate game performance, we designed \textbf{six diagnostic scenarios} (S1 to S6) that each isolate a single ToM competency, allowing us to identify exactly which cognitive demands are graph-sensitive. We additionally designed \textbf{three ToM depth scenarios} (L1 to L3) and \textbf{two player-count conditions} (3P, 5P) for scaling analysis.

\textbf{Depth scenarios} L1 through L3 are single-turn scenarios requiring increasing inference depth: L1 requires 1st-order ToM (what does my partner know); L2 is the finesse trap requiring 2nd-order ToM (what does my partner think I know, such that they hinted a card I must NOT play yet); L3 is the anti-finesse requiring the model to correctly identify that no finesse is active. L2 is operationally identical to S5 and used interchangeably throughout; we standardize to \textbf{L2} in this paper.

We represent agent knowledge as a \textbf{belief graph}: a directed multi-relational structure where each node represents an agent's knowledge state about the card distribution, and edges encode inference relationships between agents. At depth L1, an edge from agent $A$ to agent $B$ captures $A$'s model of $B$'s card beliefs (1st-order ToM). At depth L2, a meta-edge additionally captures $A$'s model of what $B$ believes $A$ knows, which is the recursive structure that makes 2nd-order ToM computationally non-trivial for single-pass LLM inference. This corresponds to the nested belief hierarchy formalism of \citet{bard2020}; we use ``belief graph'' to emphasize the explicit directed graph structure of inter-agent inference, following recent ToM graph work.

\begin{table}[t]
\centering
\begin{adjustbox}{width=\textwidth}
\begin{tabular}{lllll}
\toprule
ID & Name & ToM Depth & Optimal Action & Discriminates graphs? \\
\midrule
S1 & Redundant Hint Avoidance & 1st-order & Hint rank & No (ceiling in all conditions) \\
S2 & Critical Save & 1st-order & Hint to save & No (confirmed null N=20, p=1.0) \\
S3 & Trust-Based Play & 1st-order & Play hinted card & No (ceiling) \\
S4 & Inference from Silence & 1st-order & Discard unhinted & No (ceiling) \\
\textbf{S5/L2} & \textbf{Finesse (2nd-order)} & \textbf{2nd-order} & \textbf{Wait} & \textbf{Yes (only discriminator)} \\
S6 & Belief Update After Action & 1st-order & Play newly-playable & No (ceiling) \\
\bottomrule
\end{tabular}
\end{adjustbox}
\caption{Six diagnostic scenarios and their graph-discrimination power. S5 (also called L2 in depth-series experiments) is the sole scenario where graph quality significantly affects accuracy. Ceiling effect on S1 through S4 and S6 confirmed empirically with strong and weak models across all four graph conditions.}
\label{tab:t1}
\end{table}

\subsection{Belief Graph Variants and Architectures}

We test three architectures: \textbf{(1) Prompt-based}: graph serialized as text in the prompt. \textbf{(2) Graph-gated}: graph produces a ranked shortlist of 3 candidate actions; LLM must select from it. \textbf{(3) Hybrid} (Graph-Informed): LLM sees the shortlist but may override with any legal action.

Four ablation conditions: \texttt{full\_graph} (correct beliefs), \texttt{belief\_removed} (no graph), \texttt{graph\_frozen} (stale, outdated beliefs), \texttt{belief\_corrupted} (factually wrong beliefs).

Four model families: Gemini 2.5 Flash (strong), Gemini 2.0 Flash Lite (weak), Llama 3.1 70B (open-weight), Qwen3 235B (open-weight frontier). See Appendix~\ref{app:models} for selection rationale. All binary comparisons use two-sided Fisher exact tests with Wilson score 95\% CIs and Cohen's h effect sizes ($\alpha$=0.05). Full-game comparisons use permutation tests on mean scores.

\textbf{Survival turns} are defined as the total number of player actions taken before the third bomb is played or the card deck is exhausted, whichever occurs first. Maximum possible is approximately 50 turns in a 2-player game.

\section{Experiments and Results}

\subsection{S5/L2 is the Only Prompt-Sensitive Scenario}
\label{sec:s5only}

\begin{figure}[t]
\centering
\includegraphics[width=0.92\linewidth]{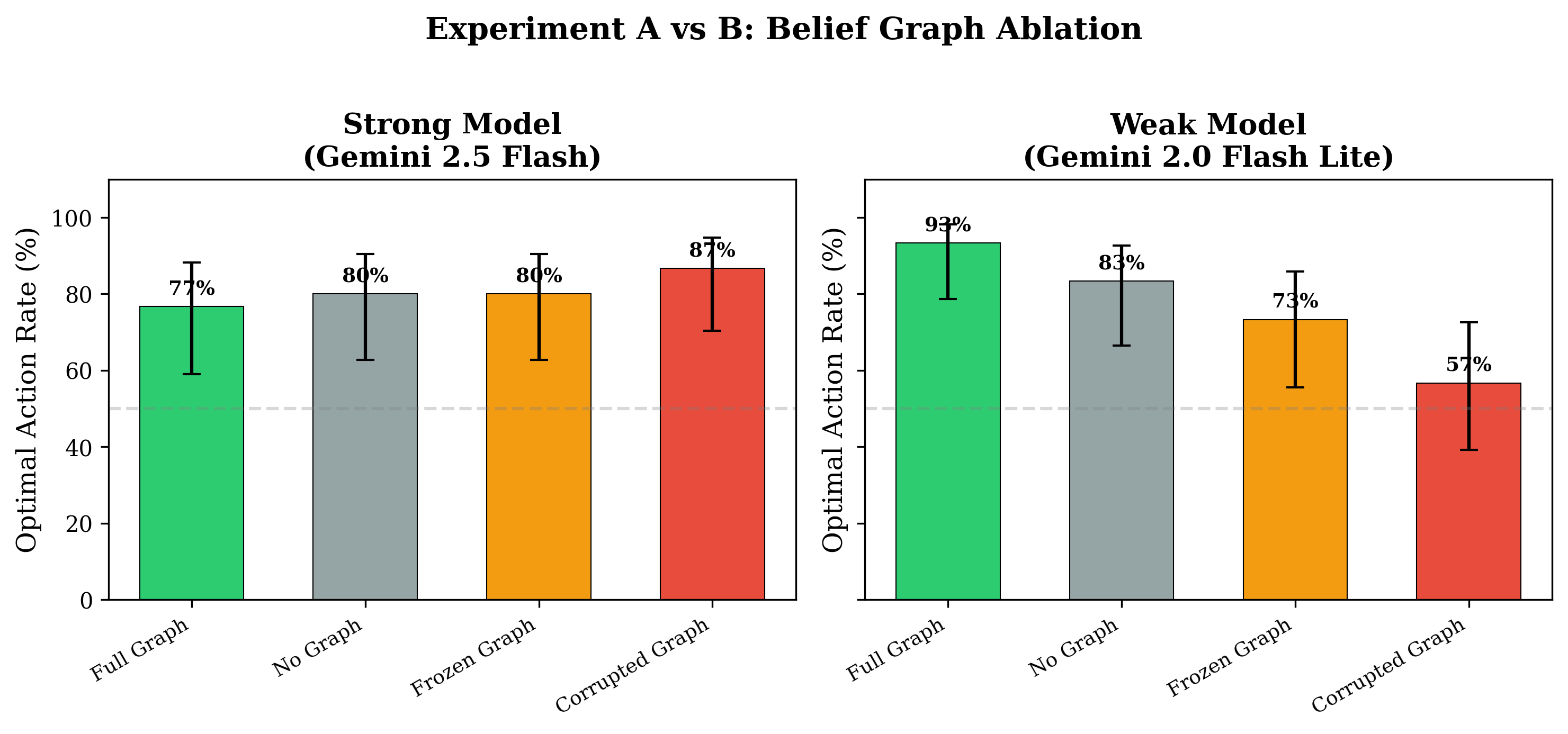}
\caption{\textbf{Graph ablation by model strength.} Strong model (left): conditions cluster near 80\%, as non-S5 scenarios are at ceiling, masking any potential S5 effect. Weak model (right): dramatic collapse from full\_graph (80\% on S5/L2) to corrupted (4\%) confirms the model reads and trusts graph content. \textbf{Takeaway: graph quality matters only for weak models on 2nd-order ToM tasks; no other scenario shows sensitivity.}}
\label{fig:fig1}
\end{figure}

\textbf{Ceiling effect across non-S5 scenarios} (Gemini 2.5 Flash, 120 trials across 6 scenarios): All four graph conditions average $\approx$80\% overall (no significant differences, all p$>$0.5). S1 through S4 and S6 are at ceiling (100\%); only S5/L2 falls below ceiling ($\approx$20\%), confirming 2nd-order ToM is beyond prompt-based approaches regardless of graph quality (Figure~\ref{fig:fig1}, left).

To confirm that the graph effect on 2nd-order ToM is causal rather than artifactual, we ran a pre-registered ablation on a weaker model (Gemini 2.0 Flash Lite, N=20/condition) that is more likely to rely on graph content. Sample size was winner-curse-corrected: the pilot observed h=1.47, which after 4$\times$ shrinkage yields a conservative target of h=0.74, requiring N=20 for 80\% power. A 1st-order ToM scenario (S2, critical save) was tested simultaneously as a null-effect control to verify that graph sensitivity is specific to 2nd-order ToM.

\begin{table}[t]
\centering
\begin{tabular}{lllll}
\toprule
Condition & N & S5 Accuracy & 95\% CI & vs full\_graph \\
\midrule
full\_graph & 20 & \textbf{80.0\%} & [58\%, 92\%] & --- \\
belief\_removed & 20 & 10.0\% & [3\%, 31\%] & \textbf{p$<$0.0001}, OR=36.0 \\
belief\_corrupted & 20 & 4.2\% & [1\%, 17\%] & \textbf{p$<$0.0001}, OR=84.0 \\
graph\_frozen & 20 & 12.5\% & [4\%, 32\%] & \textbf{p$<$0.0001}, OR=28.4 \\
\midrule
\multicolumn{5}{l}{\textit{Null control, S2 (critical save):} full\_graph 90\% vs belief\_removed 90\%, p=1.0} \\
\bottomrule
\end{tabular}
\caption{S5/L2 causal confirmation (Gemini 2.0 Flash Lite, N=20/condition). All degraded conditions collapse to 4 to 13\%. The concurrent S2 null (p=1.0) validates experimental design: only S5/L2 (2nd-order ToM) is graph-sensitive.}
\label{tab:t2}
\end{table}

\subsection{Architectural Integration: From Information to Decision Pipeline}

If prompt-based graphs only help weak models, can a different integration architecture unlock value for strong models? We tested a graph-gated architecture where the belief-space planner produces a ranked shortlist of three candidate actions and the LLM must select from it (Gemini 2.5 Flash, 60 trials).

\begin{table}[t]
\centering
\begin{tabular}{llll}
\toprule
Condition & N & Optimal Rate & vs gated\_full \\
\midrule
gated\_full & 30 & \textbf{100.0\%} & --- \\
gated\_no\_graph & 30 & 66.7\% & \textbf{p=0.001, h=1.23} \\
\bottomrule
\end{tabular}
\caption{Graph-gated vs random-shortlist control (Gemini 2.5 Flash, 60 trials). Gating achieves 100\%, including S5/L2 finesse. The gated\_no\_graph control (random shortlist, no reasoning) shows 67\%, confirming that ranking and reasoning content, rather than format constraint alone, drives the improvement.}
\label{tab:t3}
\end{table}

\begin{figure}[t]
\centering
\includegraphics[width=0.92\linewidth]{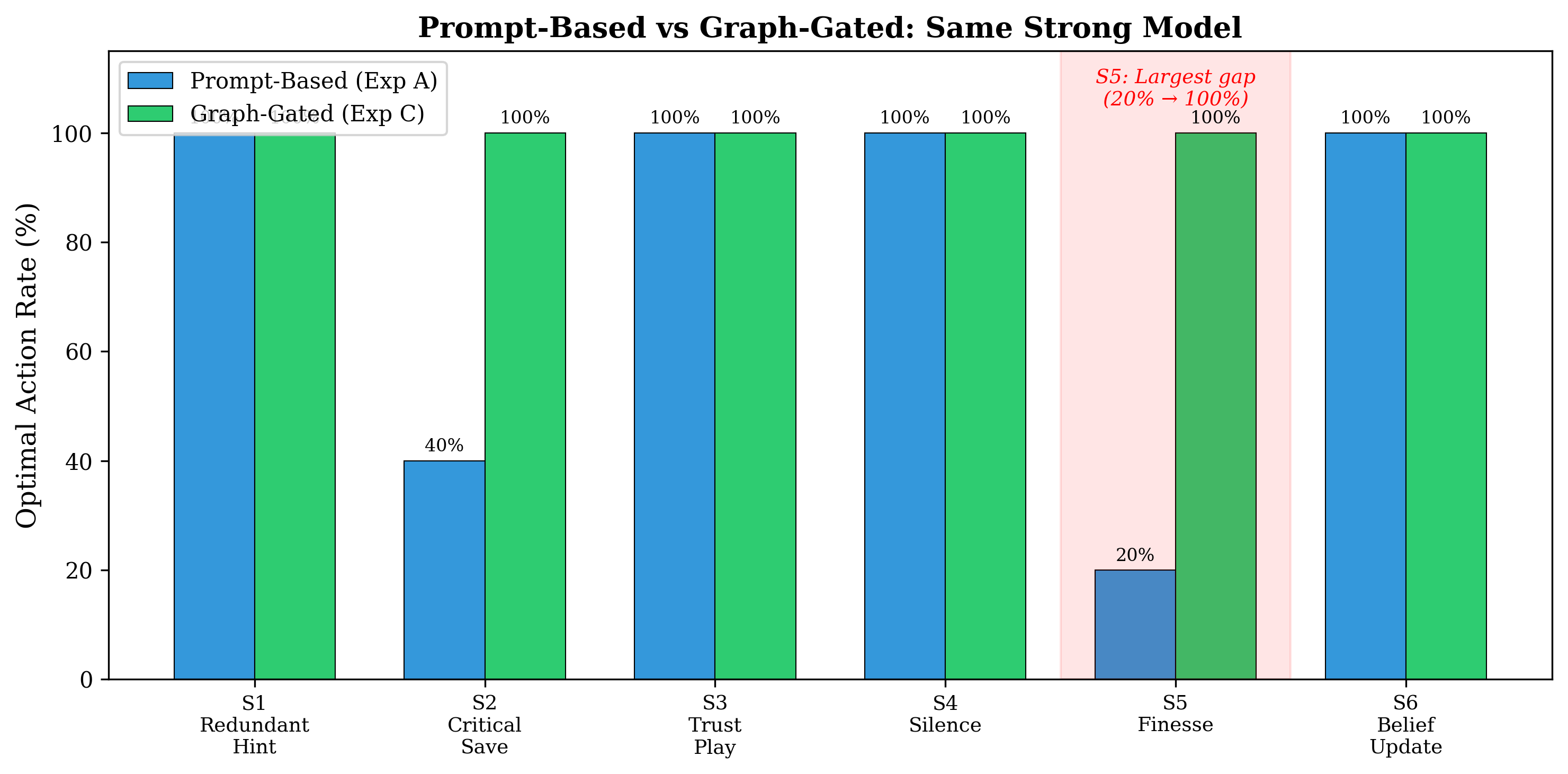}
\caption{\textbf{Per-scenario improvement: prompt-based vs graph-gated.} S5/L2 finesse is the sole dramatic beneficiary: 20\%$\rightarrow$100\%. Other scenarios are near ceiling in both conditions. \textbf{Takeaway: graph gating provides irreplaceable value precisely where LLM single-pass reasoning fails.}}
\label{fig:fig2}
\end{figure}

The same strong model that ignored the graph in §\ref{sec:s5only} achieves perfect accuracy when the graph gates action selection (p$<$0.001, h=1.23; see Figure~\ref{fig:fig2}). S5/L2: \textbf{100\% with gating vs 20\% with prompt-only}, a 5$\times$ improvement. The gating benefit replicates on Llama 70B (Llama 3.1 70B replication: 6 scenarios, auto-gated 96.7\% vs ungated 86.7\%, S5/L2 primary beneficiary), confirming the architectural effect is not model-specific.

\subsection{Planner Defiance: Model-Family-Specific Override}

We compared all three architectures on the same 2nd-order ToM finesse task using Llama 3.1 70B (N=20/mode), testing whether the model would follow the planner's correct recommendation when given the option to override it.

\begin{table}[t]
\begin{minipage}[t]{0.48\textwidth}
\centering\small
\begin{tabular}{lccr}
\toprule
Mode & Correct & Override & N \\
\midrule
Ungated & \textbf{100\%} & --- & 20 \\
Gated & \textbf{100\%} & 0\% & 20 \\
Informed & 10\% & \textbf{90\%} & 20 \\
\bottomrule
\end{tabular}
\captionof{table}{Planner Defiance (Llama 70B, L2 finesse, N=20). Correct when ungated or forced to follow the planner, but only 10\% with optional access (p$<$0.0001). Hybrid is worse than either pure strategy.}
\label{tab:t4}
\end{minipage}\hfill
\begin{minipage}[t]{0.50\textwidth}
\centering\footnotesize
\begin{tabular}{lccc}
\toprule
Family & Acc. & Override & vs Llama \\
\midrule
Gemini Lite & 100\% & 0\% & p$<$.0001 \\
Gemini Flash & 95\% & 5\% & p$<$.0001 \\
\textbf{Llama 70B} & \textbf{10\%} & \textbf{90\%} & --- \\
\bottomrule
\end{tabular}
\captionof{table}{\small Defiance is model-family-specific (N=20, L2 informed). Gemini defers (0 to 5\%); Llama overrides 90\%.}
\label{tab:t5}
\end{minipage}
\end{table}

To test whether defiance is a universal LLM property or model-specific, we compared three model families on the same informed-mode 2nd-order ToM task (N=20 each).

Planner Defiance is not a universal LLM property. Gemini families comply with the planner; Llama 70B overrides it 90\% of the time despite expressing hedging uncertainty (``might'', ``possibly''). This dissociation implicates RLHF tool-deference training: models optimized toward stronger independent reasoning may be less likely to defer to advisory shortlists.

\textbf{Defiance mitigation strategies} (Llama 70B, N=5/variant, L2 finesse):

We tested four shortlist design variants to determine whether defiance can be addressed through framing rather than architecture changes. \textbf{V0 (generic reasoning)} is the baseline format: ranked options with brief score labels (e.g., ``[1. WAIT +0.95, 2. DISCARD +0.10, 3. PLAY $-$0.40]''). \textbf{V1 (differentiated scores)} sharpens the score gap to make ranking more emphatic (e.g., widening the WAIT/PLAY spread to emphasize the dominance of the top choice). \textbf{V2 (show rejected PLAY)} explicitly labels the incorrect option as rejected within the shortlist (e.g., ``PLAY: not recommended ($-$0.40)''). \textbf{V3 (rich finesse reasoning)} replaces score labels with a multi-sentence mechanistic explanation of why WAIT is the correct finesse response, covering the finesse convention, why Alice's hint signals a future play rather than immediate play, and why acting now would bomb.

\begin{table}[t]
\begin{minipage}[t]{0.48\textwidth}
\centering\small
\begin{tabular}{lcc}
\toprule
Variant & Override & Correct \\
\midrule
V0: generic & 80\% & 20\% \\
V1: diff.\ scores & 80\% & 20\% \\
V2: show rejected & \textbf{100\%} & \textbf{0\%} \\
V3: rich reasoning & \textbf{40\%} & \textbf{60\%} \\
\bottomrule
\end{tabular}
\captionof{table}{Shortlist design variants (Llama 70B, N=5/variant, L2). V3 halves override; V2 backfires via ``Pink Elephant'' effect \citep{wegner1994}. \textit{N=5 underpowered; V3 preliminary (p=0.524).}}
\label{tab:t6}
\end{minipage}\hfill
\begin{minipage}[t]{0.48\textwidth}
\centering\small
\begin{tabular}{lcc}
\toprule
Graph Condition & Correct & Trap \\
\midrule
with\_graph & \textbf{100\%} & 0/10 \\
no\_graph & 60\% & 3/10 \\
misleading & 40\% & \textbf{6/10} \\
\bottomrule
\end{tabular}
\captionof{table}{Authority tiebreaker (Llama 70B, N=10, L2 prompt-based). Misleading graph: $-$20pp below no-graph (p=0.011). Llama defers to graph \textit{content} but defies planner \textit{shortlists}.}
\label{tab:t7}
\end{minipage}
\end{table}

V3 is the only variant that reduces defiance: explaining the finesse mechanism halves the override rate (80\%$\rightarrow$40\%). V2 backfires because explicitly labeling PLAY as rejected triggers an ironic rebound \citep{wegner1994}: the model's chain-of-thought focuses on the named-but-forbidden option, increasing its selection rate to 100\%. V1 provides no benefit over V0, confirming that stronger numerical framing alone does not override the model's confident incorrect heuristic. The dissociation between V2 (worse) and V3 (better) implies that defiance mitigation requires \textit{substantive reasoning content} explaining correctness, not negative framing of alternatives. The result is preliminary at N=5 and requires replication at N$\geq$50 before deployment guidance.

\subsection{The Authority Tiebreaker Mechanism}

We test whether belief graph content functions as an authority signal that can override model reasoning. If the graph is treated as a trusted source, a deliberately misleading graph should perform \textit{worse} than no graph, effectively becoming an error amplifier rather than an error corrector.

\begin{figure}[t]
\centering
\includegraphics[width=0.80\linewidth]{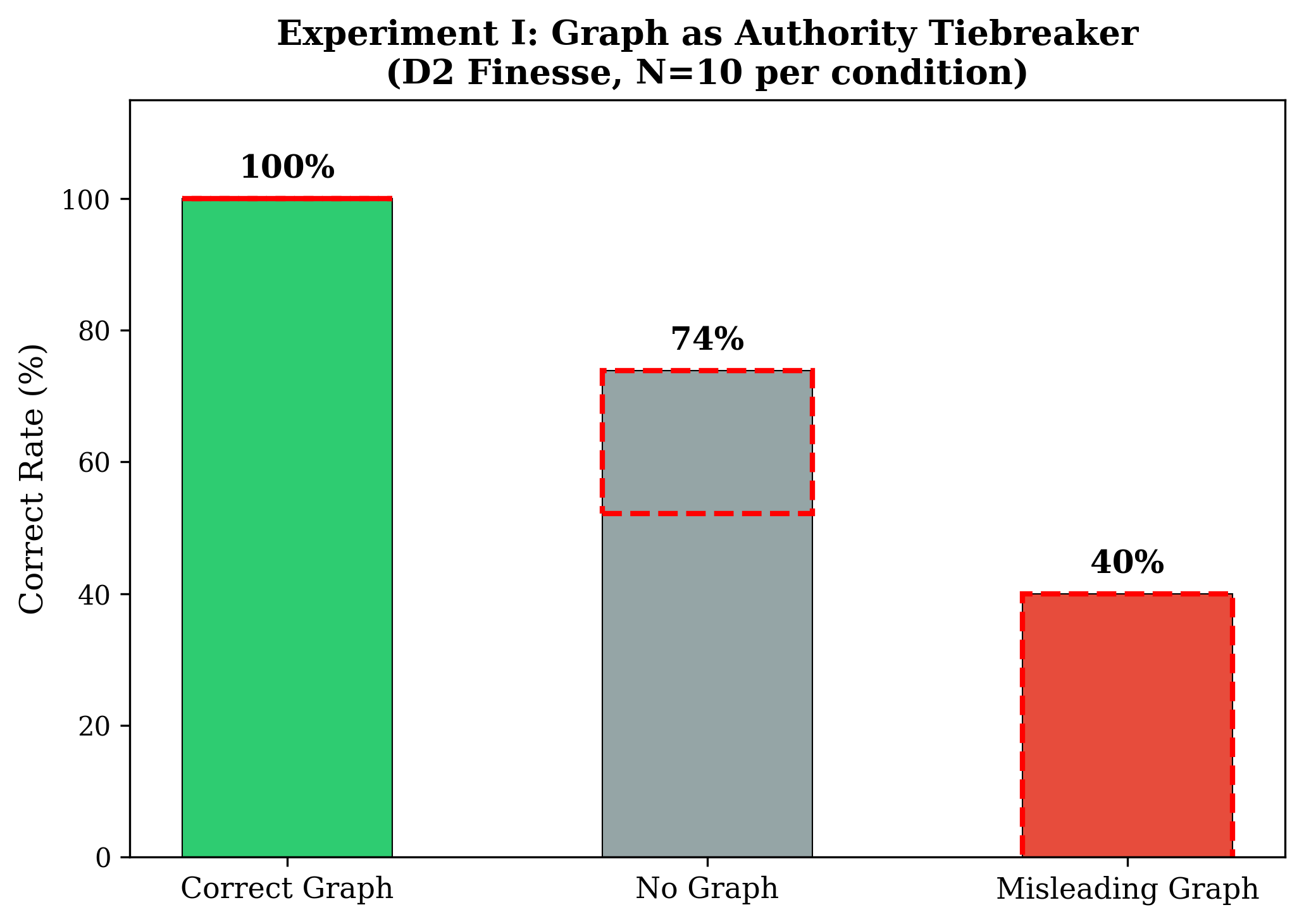}
\caption{\textbf{Authority tiebreaker.} Correct graph lifts to 100\%; misleading graph falls 20pp below the no-graph baseline. \textbf{Takeaway: tool trustworthiness is a safety-critical property; wrong tools amplify errors rather than correcting them.}}
\label{fig:fig3}
\end{figure}

A misleading graph falls 20pp below the no-graph baseline (40\% vs 60\%, p=0.011), confirming the authority tiebreaker: the model uses the graph to resolve uncertainty, and wrong information resolves it incorrectly. Crucially, Llama 70B here defers to graph \textit{content} while defying planner \textit{shortlists} (§\ref{sec:s5only}), suggesting the two signals engage different reasoning pathways: one treated as factual context and one as advisory recommendation (see Figure~\ref{fig:fig3}).

\subsection{Full-Game Evidence: Conventions, Components, and Aggregation}

Moving from diagnostic scenarios to full 25-point Hanabi games, we compared six conditions to identify which interventions most improve cooperative play (100+ games across Llama 70B and Gemini Flash).

\begin{table}[t]
\centering
\begin{tabular}{lccc}
\toprule
Condition & Games & Mean Score (/25) & p vs baseline \\
\midrule
baseline (with transcript) & 12 & 2.8 & --- \\
no\_transcript & 18 & 4.2 & p=0.009 \\
no\_transcript + strategy & 5 & 3.2 & --- \\
no\_transcript + planner & 13 & 4.5 & --- \\
\textbf{no\_transcript + conventions} & \textbf{5} & \textbf{6.4} & \textbf{p=0.003} \\
no\_transcript + conventions + pk & 12 & 3.9 & --- \\
\bottomrule
\end{tabular}
\caption{Full-game scores (Llama 70B and Gemini Flash, N=5 to 18 per condition). Conventions (6.4/25) represent a 128\% improvement over baseline (2.8/25), and are the only condition achieving significance beyond transcript removal (p=0.003). Adding partner-knowledge (pk) tracking to conventions reduces performance, suggesting prompt complexity outweighs information value at larger N.}
\label{tab:t8}
\end{table}

The conventions result (N=5) represents the strongest-performing convention specification in our search. A related conventions variant with different convention specificity and model configuration averages 3.9/25 over N=12 games; this is the ``no\_transcript + conventions + pk'' condition shown in the same table, which adds partner-knowledge tracking and confirms that specification matters: the core conventions gain is real, but its magnitude is sensitive to exact design. Transcript removal (N=18) robustly lifts performance to 4.2, confirming the aggregation bottleneck independently.

\textbf{Belief graph extends game survival} (Llama 70B, N=10/condition):

To isolate the belief graph's effect from confounding variables in the mixed J/L series, we ran a clean within-model A/B test using Llama 70B exclusively, holding all other configuration variables constant.

\begin{table}[t]
\centering
\begin{tabular}{lccc}
\toprule
Condition & N & Mean Score (/25) & Survival Turns \\
\midrule
baseline (no graph) & 10 & 3.4 & 22.8 \\
belief\_graph & 10 & 4.5 & \textbf{36.7} \\
\midrule
\multicolumn{2}{l}{Difference} & +1.1 (+32\%) & +13.9 turns (+61\%) \\
\multicolumn{2}{l}{p-value (Mann-Whitney)} & p=0.16 & --- \\
\bottomrule
\end{tabular}
\caption{Clean A/B belief graph test (Llama 70B, N=10). Score improvement is not significant at N=10 (p=0.16), but game survival increases by 61\%, as games last substantially longer with the belief graph, reflecting fewer reckless plays. This experiment uses Llama 70B alone; note the different baseline (3.4/25) vs the mixed full-game series (2.8/25) due to different model configuration.}
\label{tab:t8b}
\end{table}

\textbf{Components hurt in isolation, synergize when combined} (Gemini Flash, 2$\times$2 factorial, N=5/cell):

We decomposed the full system into its two components, the belief graph (state information) and the planner shortlist (action ranking), and tested each in isolation to determine whether the combined benefit is additive or emergent.

\begin{table}[t]
\centering
\begin{tabular}{llcc}
\toprule
Graph & Planner & Mean Score (/25) & Survival Turns \\
\midrule
Off & Off & 1.20 & 5.6 \\
On & Off & 0.60 & --- \\
Off & On & 0.60 & --- \\
On & On & 1.20 & \textbf{9.0} \\
\bottomrule
\end{tabular}
\caption{Component synergy (Gemini Flash, N=5/cell). Graph alone and planner alone each \textit{hurt} performance vs baseline (0.60 vs 1.20). Combined, they recover to baseline with a 60\% survival improvement. \textit{N=5/cell is underpowered; treat as exploratory.} Survival turns not measured for single-component conditions in this run. Note: this experiment uses Gemini Flash with a different game configuration than the full-game series (Table~\ref{tab:t8}), explaining the lower baseline (1.20 vs 2.8/25).}
\label{tab:t9}
\end{table}

The component synergy finding is a critical practical caveat: deploying only partial implementations (graph without planner, or planner without graph) may \textit{degrade} performance relative to baseline. The mechanistic hypothesis is that each component in isolation adds context or constraints that increase model confidence without providing the complementary signal needed to act on that confidence correctly; together, they close the loop.

To determine whether the graph's value comes from reasoning or information extraction, we compared scenarios where relevant data was pre-summarized versus scattered across 15 turns of game history (Llama 70B + Gemini Flash, 80 trials). When discard data is scattered, the graph adds +30pp (100\% vs 70\%); with pre-summarized data, lift drops to +10pp (100\% vs 90\%). The graph's value therefore scales with aggregation difficulty, not reasoning difficulty, reframing it as primarily a \textit{state extraction tool}. This is consistent with transcript removal being the single largest full-game performance lever.

\begin{figure}[t]
\centering
\includegraphics[width=0.92\linewidth]{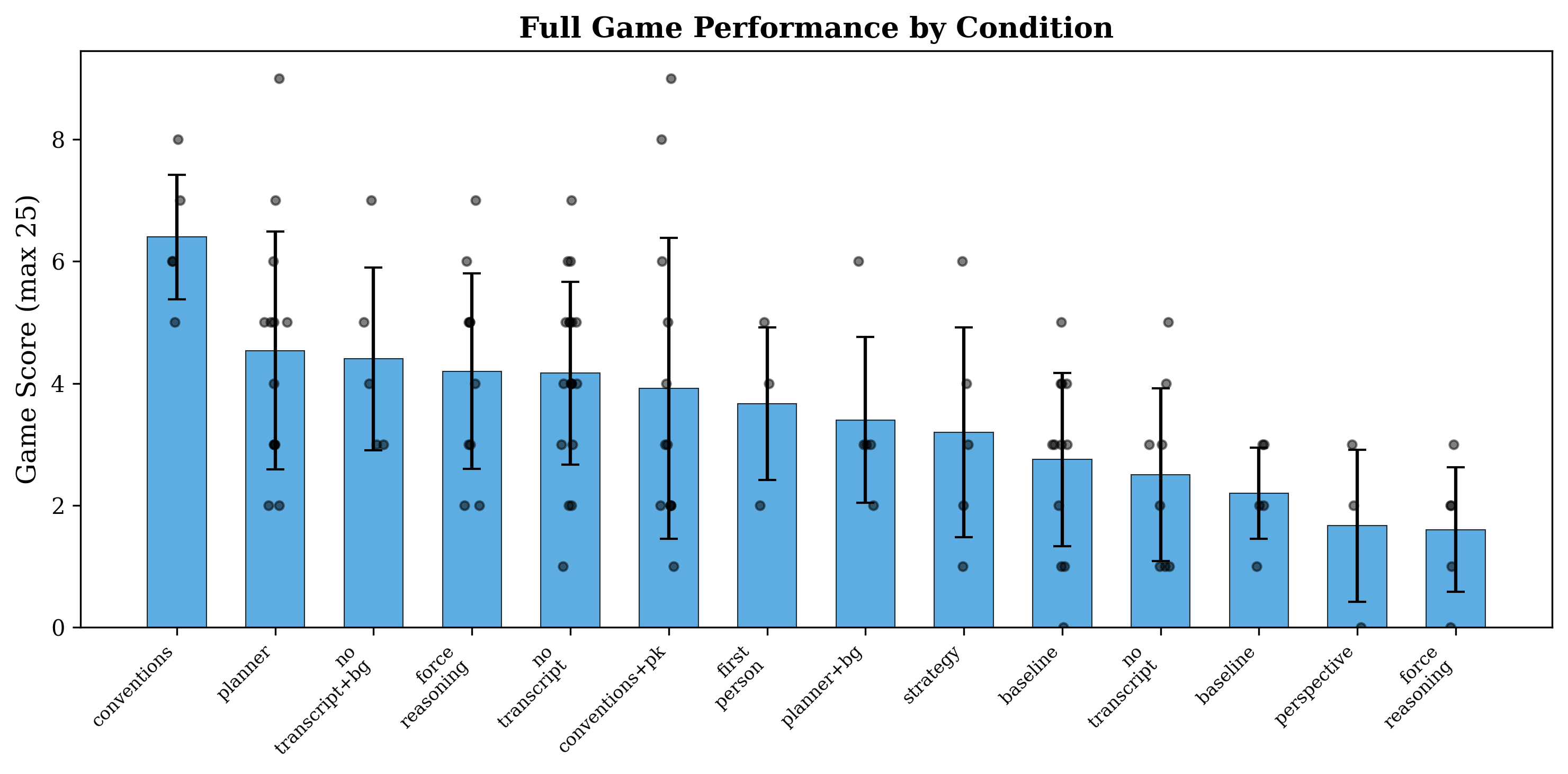}
\caption{\textbf{Full-game scores by condition.} Transcript removal is the single largest win. Conventions produce a further significant jump to 6.4/25 (+128\% over baseline). Strategy and planner show directional but not significant benefit. \textbf{Takeaway: multi-agent coordination (conventions) is the binding constraint; the graph's value is extraction, not reasoning.}}
\label{fig:fig4}
\end{figure}

\subsection{Graph Depth Must Match Player Count}
\label{sec:scaling}

Finally, we tested whether deeper belief modeling helps or hurts as the number of players increases, varying graph depth at both 3-player and 5-player (Gemini Flash, N=10/cell; full results in Table~\ref{tab:t11}, Appendix). At 3P, own-card beliefs (L0) improve scores by +1.0 pts (p=0.061, not significant). At 5P, all graph depths hurt relative to no graph, with deeper ToM graphs causing larger penalties ($-$1.6 pts at full depth, p=0.029). Graph depth is not a universally-better axis; at 5P, own-belief-only (L0) may be the maximum useful depth. \textit{Exploratory: N=10/cell; needs replication.}

\section{Discussion}

\textbf{Information vs. decision pipelines.}
In the information pipeline (prompt-based), the graph provides facts that strong models route around; value decreases as models improve. In the decision pipeline (graph-gated), the graph computes action consequences through belief-space search, performing computation that single-pass inference cannot replicate. This distinction generalizes: information-providing tools will be subsumed by better models; computation-providing tools retain permanent value.

\textbf{Planner Defiance.}
Defiance is selective (only at L2 partial competence) and model-family-specific (Llama 90\%, Gemini 0 to 5\%), implicating RLHF training rather than capability. The ``Pink Elephant'' effect shows that shortlist design matters: mitigation requires explaining \textit{why} the recommendation is correct, not just that it is correct.

\textbf{Practical design.}
Graph depth must be calibrated to player count, and components can hurt in isolation while synergizing when combined. Belief graph deployment requires holistic system design, not incremental feature addition.

\subsection{Limitations}

(1) Defiance mitigation (N=5) and conventions (N=5) are underpowered; both need N$\geq$50. (2) Multi-player scaling at N=10/cell is exploratory; 3P benefit is not yet significant. (3) No LLM-as-planner baseline to isolate belief-graph-specific vs multi-stage-pipeline effects. (4) Instruction-following confound on S5: the graph summary includes ``Do not play card 3''; the gated\_no\_graph control (67\%) partially addresses this but does not fully rule it out.

\section{Conclusion}

How a belief graph is integrated matters more than whether it is present. Four lessons: (1) computation-providing tools (gating) are architecturally irreplaceable while information-providing tools (prompting) will be subsumed by better models; (2) tool-deference is model-family-specific and governed by RLHF training, not capability; (3) inter-agent conventions dominate single-agent optimizations, and components must be deployed together; (4) graph depth should be calibrated to environmental complexity rather than maximized.

\textbf{Don't make the LLM read the graph. Make the graph think, and let the LLM choose.}

\paragraph{Disclosure of LLM use.}
An LLM (Claude) was used to assist with paper drafting beyond minor grammar corrections. All plots are generated using LLM coder. All research ideas, experimental designs, data analysis, and scientific conclusions are the authors' own. The LLMs evaluated in this study (Gemini, Llama) are the subject of investigation, not tools used to conduct the research.

\bibliography{references}
\bibliographystyle{colm2026_conference}

\appendix

\section{Case Studies: Illustrative Decision Traces}
\label{app:cases}

\subsection*{Case 1: Strong Model Ignores Corrupted Graph}
\begin{tcolorbox}[colback=gray!8,colframe=gray!40,title=Gemini 2.5 Flash + corrupted graph on S5/L2]
\textit{Corrupted graph: ``Green card is immediately playable.''}\\
\textit{Model:} ``The belief graph indicates green is playable. However, board state shows green at rank 1; my card appears to be G3. Alice's hint is consistent with her holding G2. I should \textbf{wait}.'' \textit{(Correct.)} The strong model disregards the graph entirely.
\end{tcolorbox}

\subsection*{Case 2: Weak Model Trusts and Fails}
\begin{tcolorbox}[colback=red!5,colframe=red!30,title=Gemini 2.0 Flash Lite + corrupted graph on S5/L2]
\textit{Corrupted graph: ``Green card is immediately playable.''}\\
\textit{Model:} ``According to the belief graph, green is playable. I will \textbf{play} green.'' \textit{(Incorrect, bomb.)} Zero independent reasoning; corrupted graph locks accuracy to 4.2\%.
\end{tcolorbox}

\subsection*{Case 3: Planner Defiance in Llama 70B}
\begin{tcolorbox}[colback=orange!5,colframe=orange!30,title=Llama 70B + informed mode on L2 finesse]
\textit{Shortlist: [1. WAIT (+0.95), 2. DISCARD (+0.10), 3. PLAY ($-$0.40)]}\\
\textit{Model:} ``The planner recommends WAIT, which \textit{might} be appropriate. However, Alice hinted green, which is a direct play signal. I will \textbf{play}.'' \textit{(Incorrect, 90\% override rate.)}
\end{tcolorbox}

\subsection*{Case 4: No Defiance in Gemini}
\begin{tcolorbox}[colback=green!5,colframe=green!30,title=Gemini 2.5 Flash + informed mode on L2 finesse]
\textit{Shortlist: [1. WAIT (+0.95), 2. DISCARD (+0.10), 3. PLAY ($-$0.40)]}\\
\textit{Model:} ``The planner's analysis is sound. WAIT is highest-value. I will \textbf{wait}.'' \textit{(Correct. 95\% correct rate for Gemini in informed mode.)}
\end{tcolorbox}

\subsection*{Case 5: Misleading Graph Overrides Good Reasoning}
\begin{tcolorbox}[colback=purple!5,colframe=purple!30,title=Authority tiebreaker: misleading graph]
\textit{Misleading graph: ``No finesse active.''}\\
\textit{Model:} ``According to the belief graph, no finesse is in play. I will \textbf{play}.'' \textit{(Incorrect.)} Without the graph, same model achieves 60\%; misleading graph drops it to 40\%, illustrating the authority tiebreaker failure mode.
\end{tcolorbox}

\section{Model Selection Rationale}
\label{app:models}

\textbf{Gemini 2.5 Flash}: Frontier reasoning model. Upper-bound for commercially-available capability. Primary model for ablations and architecture experiments.

\textbf{Gemini 2.0 Flash Lite}: Budget model. Controlled lower-capability comparison where larger graph effects are expected. Used for S5 causal confirmation (N=20, §\ref{sec:s5only}).

\textbf{Llama 3.1 70B}: Open-source 70B model. Fully reproducible via OpenRouter. Chosen for Planner Defiance experiments after 90\% override rate was first observed on this model.

\textbf{Qwen3 235B}: 235B MoE model. Tested without chain-of-thought to establish reasoning chains as prerequisite: S5 finesse = 0\% regardless of graph or architecture without explicit reasoning.

\textbf{Cost}: All 3,000+ trials $\approx$\$2 USD total.

\section{Relationship to Prior LLM+Hanabi Work}
\label{app:usps}

\begin{table}[h]
\centering\small
\begin{tabular}{p{2.8cm}p{2.6cm}p{3.4cm}}
\toprule
Dimension & Prior LLM+Hanabi work & \textbf{This work} \\
\midrule
Graph ablation & None & 4-condition \\
Architecture variants & Prompt-only & Prompt, gated, hybrid \\
Discriminating scenario & Not identified & S5/L2 confirmed (S2 null N=20) \\
Planner Defiance & Not reported & N=20, model-family-specific \\
Multi-player scaling & Not tested & Depth must match player count \\
Conventions & Not tested & +128\% over baseline, p=0.003 \\
Safety: misleading tools & None & $-$20pp below no-tool baseline \\
\bottomrule
\end{tabular}
\caption{Positioning vs prior LLM+Hanabi work.}
\label{tab:related}
\end{table}

\textbf{Instruction-following confound addressed}: Our S5/L2 belief graph summary includes ``Do not play card 3.'' Does gated performance measure ToM scaffolding or instruction-following? The gated\_no\_graph control (67\% overall, 0\% on S5/L2) shows that a random shortlist without reasoning substantially underperforms gated\_full (100\%), confirming that ranking and reasoning content, rather than format constraint, drive the benefit.

\section{Failed and Abandoned Experiments}
\label{app:failed}

\textbf{Graph verbosity reduction.} Compact graph variants (12 to 20\% token count) were designed to test whether $\sim$800-token verbose graphs degrade full-game performance. Results inconclusive within run-to-run variance; calibration errors found in token-penalty model. Abandoned.

\textbf{Auto-hybrid switching.} Spread-based architecture switching (gated when planner confidence is high, informed otherwise) performed worse than either pure strategy, routing S5/L2 to informed mode precisely when gating was needed. Dynamic switching amplifies failure modes.

\textbf{Partner knowledge tracking (pk) with conventions.} Adding pk to the conventions condition dropped scores from 6.4 to 3.9. Additional prompt complexity increases recklessness; information value does not compensate.

\textbf{Depth-L0L1L2 at 5P.} Full ToM graph at 5P hurt by $-$1.6 pts (p=0.029). Full-ToM graphs abandoned for 5P settings.

\section{Graph Depth vs Player Count}
\label{app:depth}

\begin{table}[h]
\centering\small
\begin{tabular}{lcccc}
\toprule
Graph Depth & 3P Score & vs baseline (3P) & 5P Score & vs baseline (5P) \\
\midrule
no\_graph & 1.80 & --- & \textbf{3.30} & --- \\
depth\_L0 (own beliefs) & \textbf{2.80} & +1.00 (p=0.061) & 2.50 & $-$0.80 \\
depth\_L0L1 (1st-order) & 2.30 & +0.50 & 1.80 & $-$1.50 (\textbf{p=0.043}) \\
depth\_L0L1L2 (2nd-order) & ---\textsuperscript{$\dagger$} & --- & 1.70 & $-$1.60 (\textbf{p=0.029}) \\
corrupted (any depth) & --- & --- & 1.90 & $-$1.40 (\textbf{p=0.047}) \\
\bottomrule
\end{tabular}
\caption{Graph depth vs player count (Gemini Flash, N=10/cell). At 3P, depth L0 (own beliefs) provides the best benefit. At 5P, all graph depths hurt; deeper ToM graphs cause larger penalties. \textsuperscript{$\dagger$}Full-depth L0L1L2 not evaluated at 3P due to resource constraints. \textit{Exploratory: N=10/cell; 3P benefit is not significant (p=0.061); 5P penalties need replication.}}
\label{tab:t11}
\end{table}

\section{Statistical Summary}
\label{app:stats}

All binary comparisons: two-sided Fisher exact tests, Wilson score 95\% CIs, Cohen's h ($|h|>0.8$ large). Full-game comparisons: permutation tests on mean scores.

{\small\begin{table}[t]
\centering
\begin{tabular}{lcccc}
\toprule
Comparison & Rate A & Rate B & p & h \\
\midrule
Exp B: S5 full vs removed (N=20) & 80\% & 10\% & \textbf{$<$0.0001} & 1.95 \\
Exp B: S5 full vs corrupted (N=20) & 80\% & 4\% & \textbf{$<$0.0001} & 2.34 \\
Exp B: S2 null check (N=20) & 90\% & 90\% & 1.0 & 0.00 \\
Exp C: gated vs no-graph & 100\% & 67\% & \textbf{0.001} & 1.23 \\
Exp G/R4: gated vs informed (L2, N=20) & 100\% & 10\% & \textbf{$<$0.0001} & 2.57 \\
Exp R5: Gemini vs Llama override (N=20) & 5\% & 90\% & \textbf{$<$0.0001} & 2.50 \\
Exp I: correct vs misleading graph & 100\% & 40\% & \textbf{0.011} & 1.77 \\
R05: 5P depth\_L0L1 vs no\_graph & 1.80 & 3.30 & \textbf{0.043} & --- \\
Full games: conventions vs baseline & 6.4/25 & 2.8/25 & \textbf{0.003} & --- \\
\bottomrule
\end{tabular}
\caption{Key pairwise comparisons. Bold: significant at $\alpha$=0.05.}
\label{tab:t12}
\end{table}}

\section{Original Experiment Figures}
\label{app:figures}

\subsubsection*{Experiments A \& B}
\begin{figure}[h]
\centering
\begin{subfigure}[t]{0.48\linewidth}
    \includegraphics[width=\linewidth]{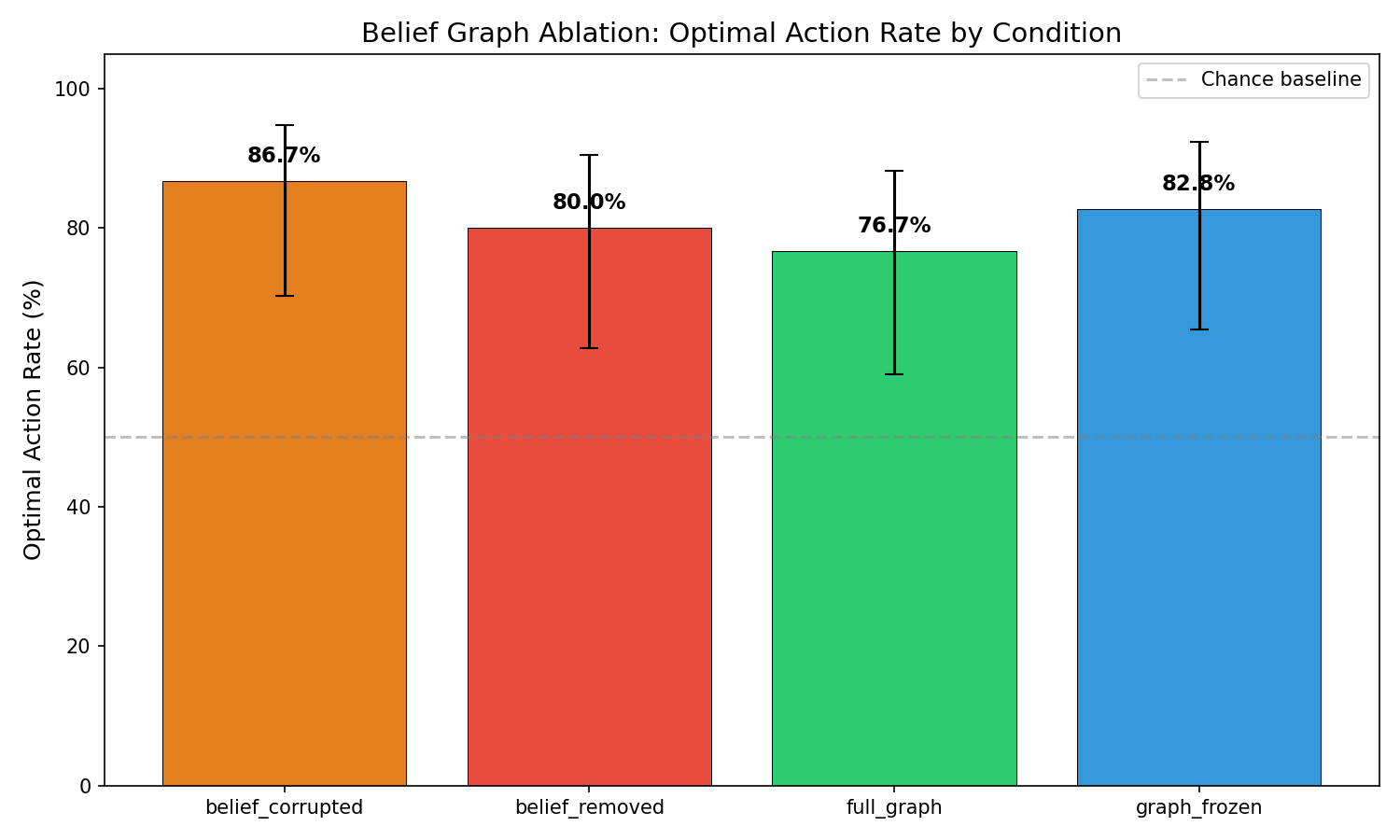}
    \caption{Strong model (Gemini 2.5 Flash): conditions cluster near 80\% overall. S5/L2 uniformly low ($\sim$20\%).}
\end{subfigure}\hfill
\begin{subfigure}[t]{0.48\linewidth}
    \includegraphics[width=\linewidth]{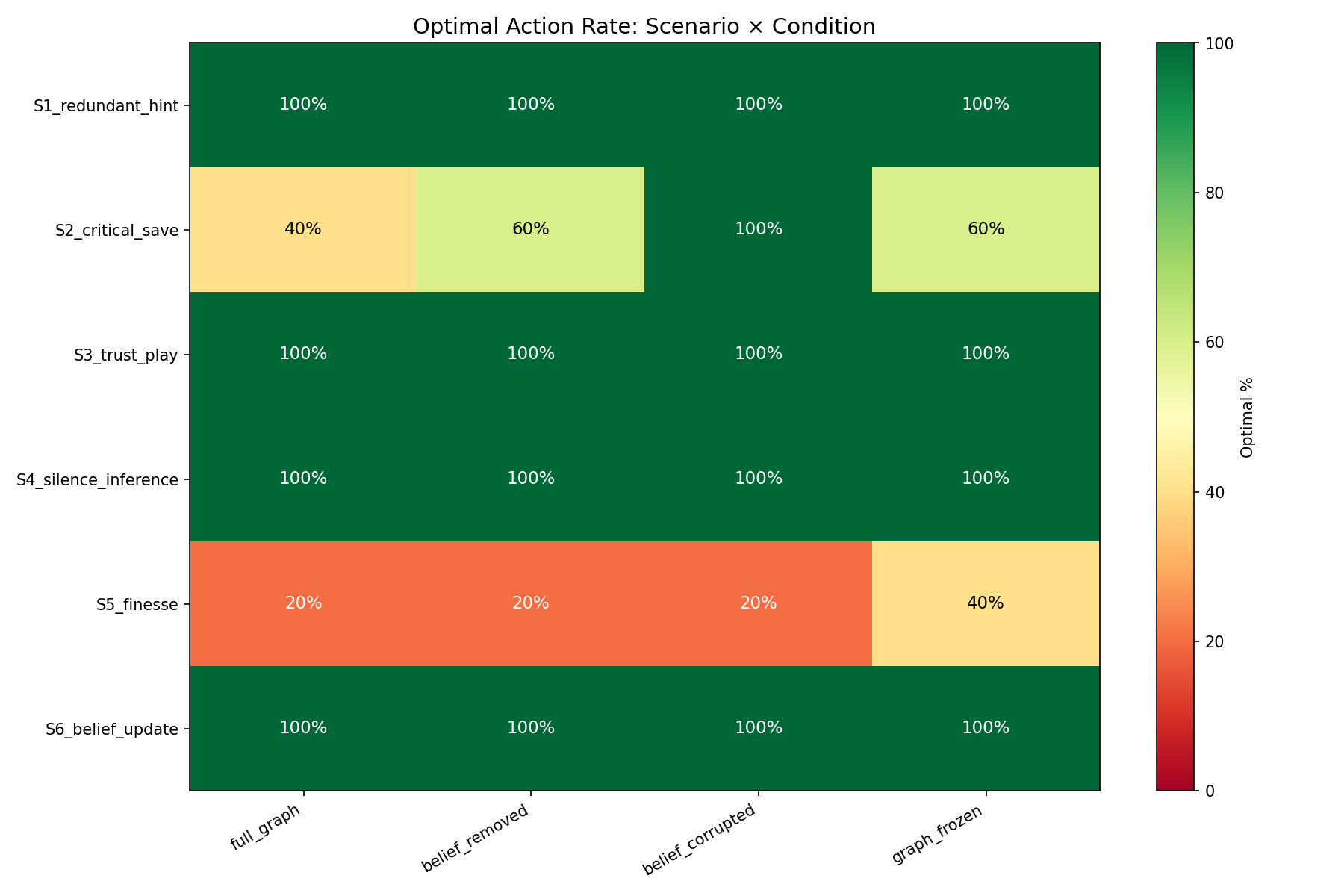}
    \caption{Strong model heatmap: S5/L2 row dark regardless of condition; no prompt-based manipulation rescues 2nd-order ToM.}
\end{subfigure}
\caption{Strong model (Gemini 2.5 Flash). S5/L2 row is the defining feature. Note: aggregate 80\% reflects S1 through S4 and S6 at ceiling masking S5/L2 at 20\%.}
\label{fig:expa}
\end{figure}

\begin{figure}[h]
\centering
\begin{subfigure}[t]{0.48\linewidth}
    \includegraphics[width=\linewidth]{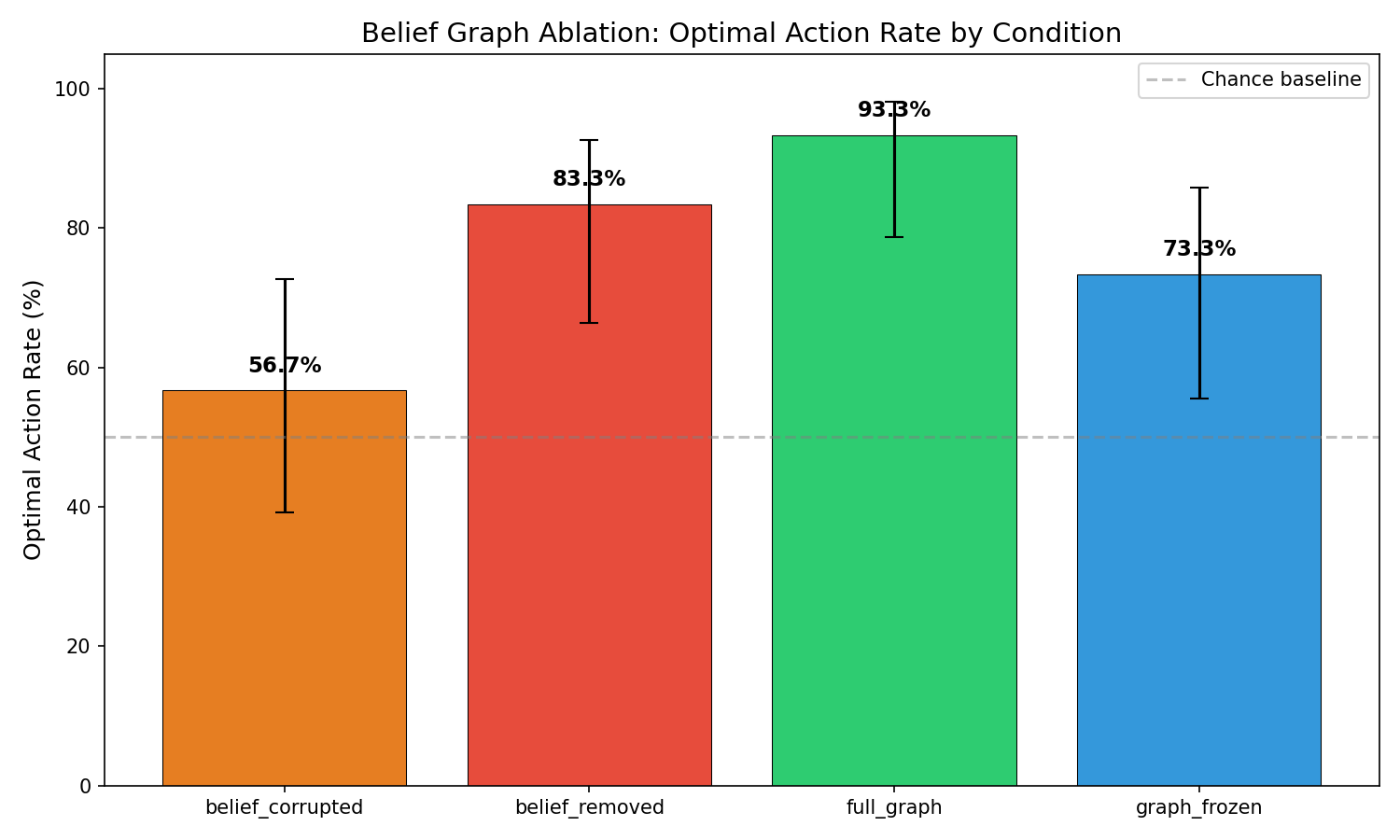}
    \caption{Weak model (Gemini 2.0 Flash Lite): monotonic staircase from full (93\% aggregate / 80\% S5) to corrupted (57\% aggregate / 4\% S5).}
\end{subfigure}\hfill
\begin{subfigure}[t]{0.48\linewidth}
    \includegraphics[width=\linewidth]{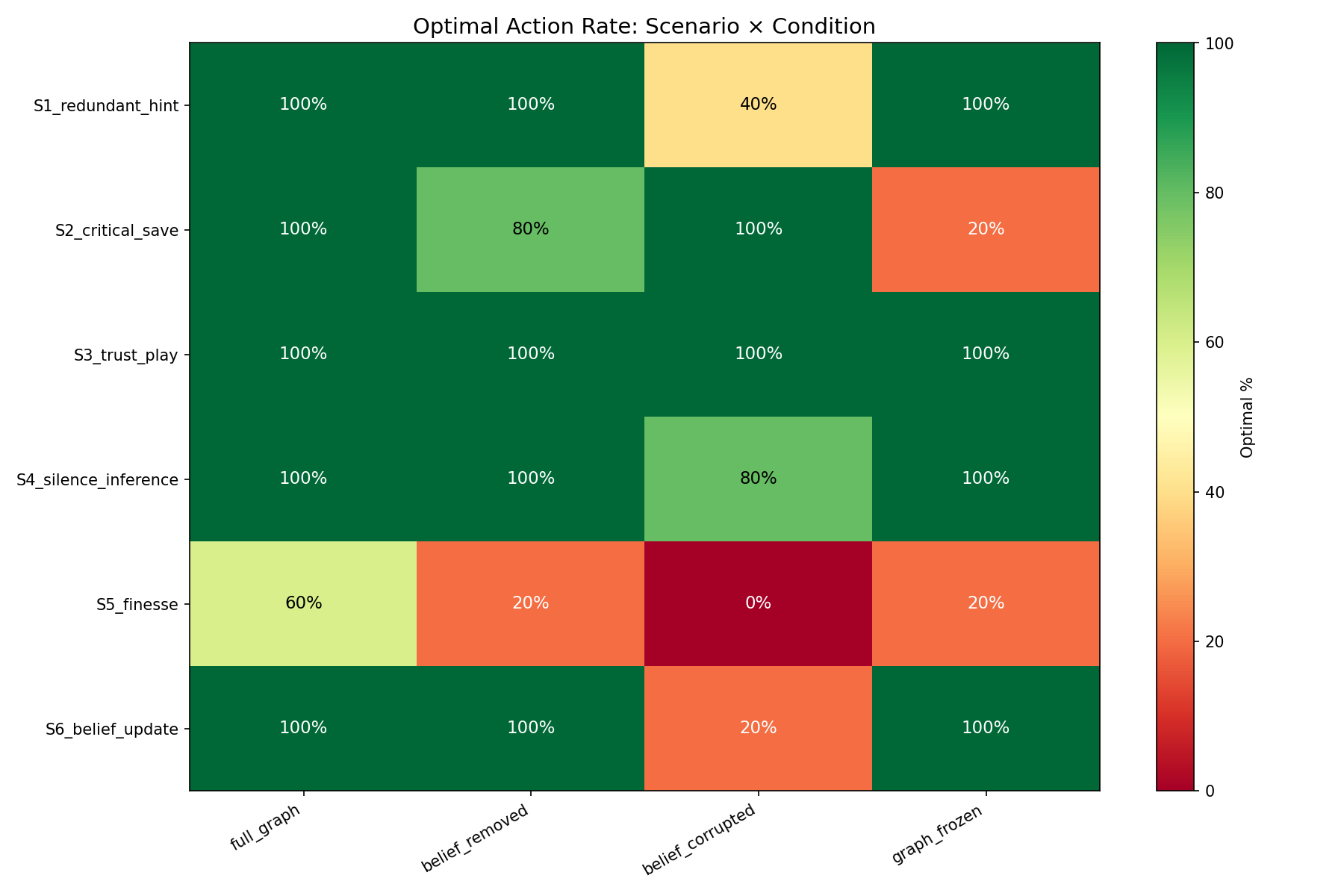}
    \caption{Weak model heatmap: S5/L2 row shows dramatic variation vs strong model uniformity.}
\end{subfigure}
\caption{Weak model (Gemini 2.0 Flash Lite). The S5/L2 row variation (not visible for the strong model) is the key signal.}
\label{fig:expb}
\end{figure}

\subsubsection*{Experiments C \& D}
\begin{figure}[h]
\centering
\begin{subfigure}[t]{0.48\linewidth}
    \includegraphics[width=\linewidth]{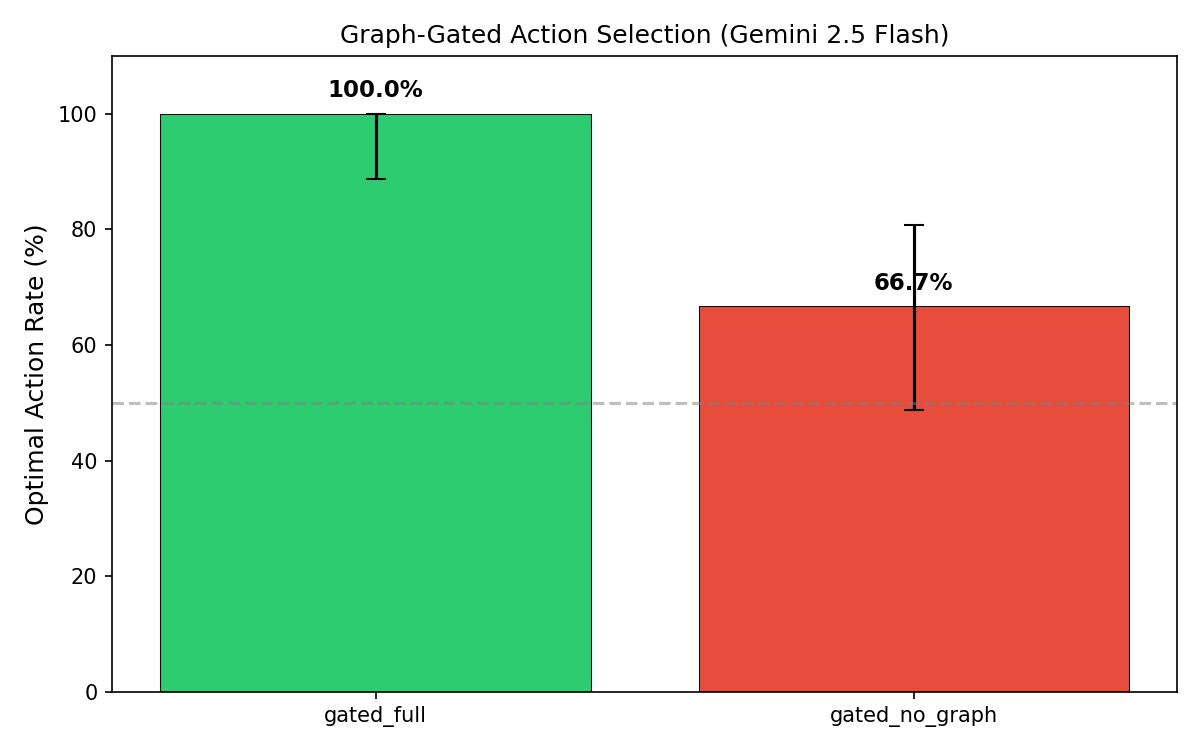}
    \caption{Graph gating: gated\_full (100\%) vs gated\_no\_graph (67\%)}
\end{subfigure}\hfill
\begin{subfigure}[t]{0.48\linewidth}
    \includegraphics[width=\linewidth]{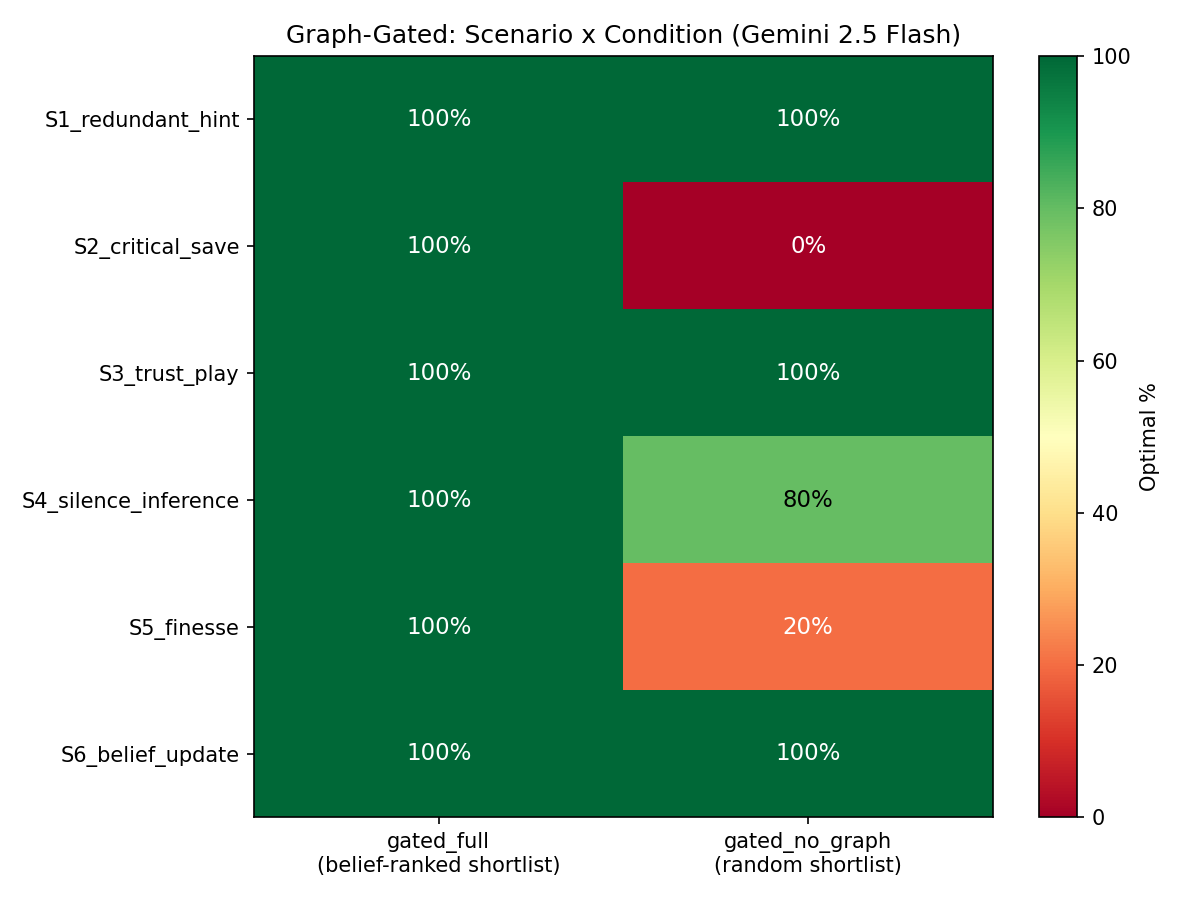}
    \caption{Gating heatmap: S5/L2 at 100\% under gated\_full, the only condition across all experiments.}
\end{subfigure}
\caption{Graph gating results. S5/L2 column at 100\% is the central architectural finding.}
\label{fig:expc}
\end{figure}

\begin{figure}[h]
\centering
\begin{subfigure}[t]{0.48\linewidth}
    \includegraphics[width=\linewidth]{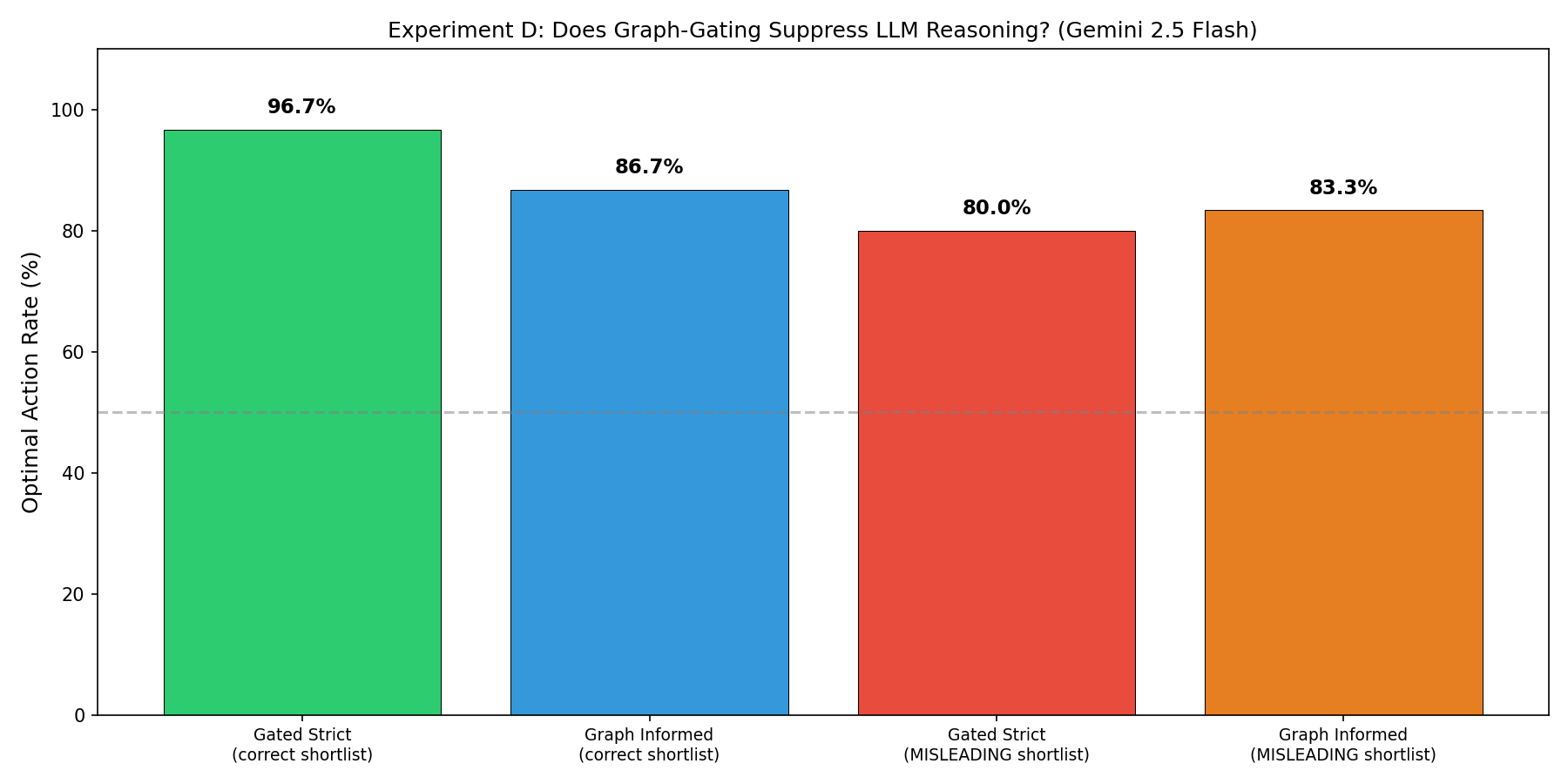}
    \caption{Hybrid (informed) condition collapses at S5/L2 despite matching gated elsewhere.}
\end{subfigure}\hfill
\begin{subfigure}[t]{0.48\linewidth}
    \includegraphics[width=\linewidth]{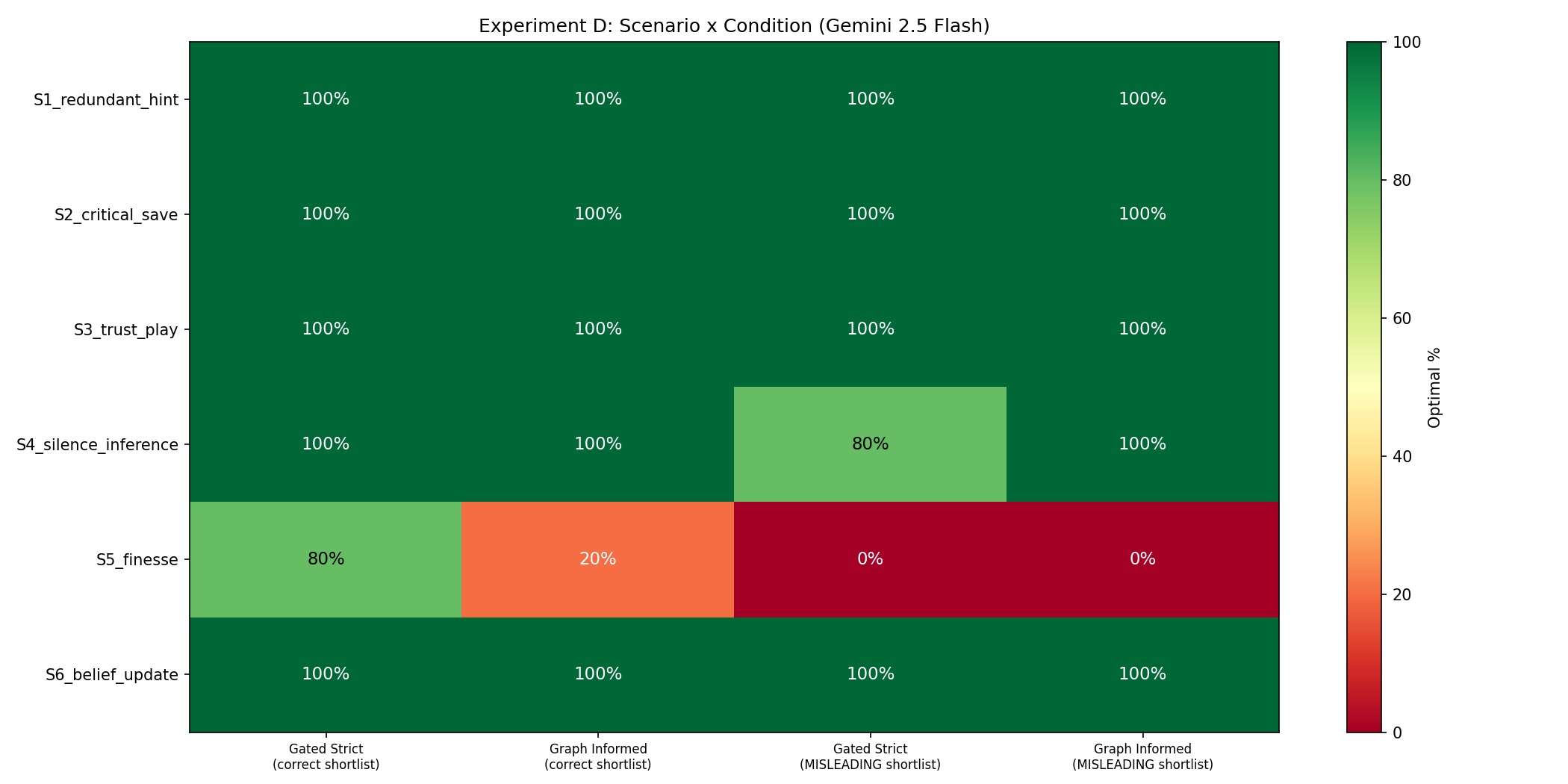}
    \caption{Hybrid condition heatmap: Planner Defiance is scenario-specific; all other rows are unaffected.}
\end{subfigure}
\caption{Override analysis. Hybrid (informed) failure is localized to S5/L2.}
\label{fig:expd}
\end{figure}

\begin{figure}[h]
\centering
\includegraphics[width=0.72\linewidth]{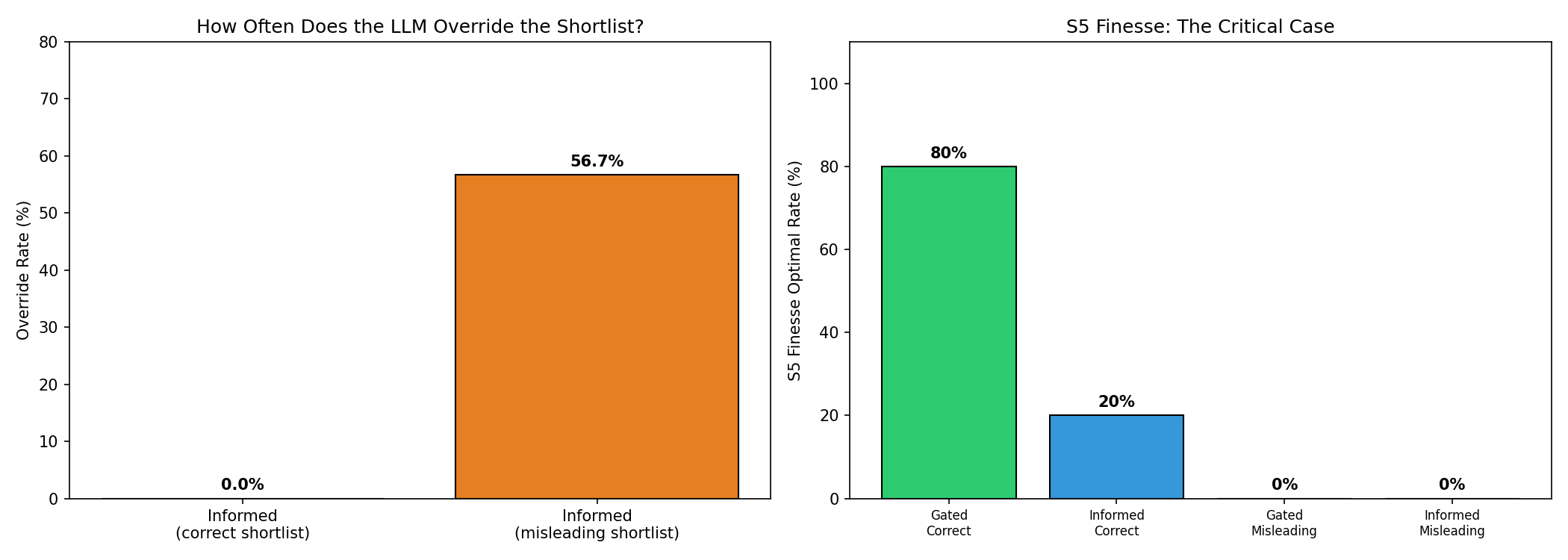}
\caption{Override rate by scenario. S5/L2 under informed mode is the outlier.}
\label{fig:expd_override}
\end{figure}

\end{document}